\documentclass[11pt]{article}
\usepackage{acl2016}
\usepackage{times}
\usepackage{url}
\usepackage{latexsym}
\usepackage{amsmath}
\usepackage{amssymb}
\usepackage{graphicx}
\usepackage{color}
\usepackage[english]{babel}
\aclfinalcopy 

\newcommand{\x}{\boldsymbol{x}}

\title{Explaining Predictions of Non-Linear Classifiers in NLP}

\author{Leila Arras$^1$, Franziska Horn$^2$, Gr{\'e}goire Montavon$^2$,\\ \textbf{ Klaus-Robert M{\"u}ller$^{2,3}$, and Wojciech Samek$^1$}\\
  $^1$Machine Learning Group, Fraunhofer Heinrich Hertz Institute, Berlin, Germany \\
  $^2$Machine Learning Group, Technische Universit\"at Berlin, Berlin, Germany \\
  $^3$Department of Brain and Cognitive Engineering, Korea University, Seoul, Korea \\
  {\tt \{leila.arras, wojciech.samek\}@hhi.fraunhofer.de}\\
  {\tt klaus-robert.mueller@tu-berlin.de}}

\date{}

\begin{document}
\maketitle
\begin{abstract}
Layer-wise relevance propagation (LRP)  is a recently proposed technique for explaining predictions of complex non-linear classifiers in terms of input variables. In this paper, we apply LRP for the first time to natural language processing (NLP). More precisely, we use it to explain the predictions of a convolutional neural network (CNN) trained on a topic categorization task. Our analysis highlights which words are relevant for a specific prediction of the CNN. We compare our technique to standard sensitivity analysis, both qualitatively and quantitatively, using a ``word deleting'' perturbation experiment, a PCA analysis, and various visualizations. All experiments validate the suitability of LRP for explaining the CNN predictions, which is also in line with results reported in recent image classification studies.\footnote{Paper will appear in the Proceedings of the 1st Workshop on Representation Learning for NLP at Association for Computational Linguistics Conference (ACL 2016).}
\end{abstract}

\section{Introduction}
Following seminal work by \newcite{Bengio} and \newcite{Collobert}, the use of deep learning models for natural language processing (NLP) applications received an increasing attention in recent years.
In parallel, initiated by the computer vision domain, there is also a trend toward understanding deep learning models through visualization techniques \cite{Erhan,Landecker,Zeiler,Simonyan,Bach,LapCVPR16} 
or through decision tree extraction \cite{Krishnan}.
Most work dedicated to understanding neural network classifiers for NLP tasks \cite{Denil2,Li} use gradient-based approaches. Recently, a technique called layer-wise relevance propagation (LRP) \cite{Bach} has been shown to produce more meaningful explanations in the context of image classifications \cite{SamArXiv15}. In this paper, we apply the same LRP technique to a NLP task, where a neural network maps a sequence of \textit{word2vec} vectors representing a text document to its category, and evaluate whether similar benefits in terms of explanation quality are observed.

In the present work we contribute by (1) applying the LRP method to the NLP domain, (2) proposing a technique for quantitative evaluation of explanation methods for NLP classifiers, and (3) qualitatively and quantitatively comparing two different explanation methods, namely LRP and a gradient-based approach, on a topic categorization task using the \textit{20Newsgroups} dataset.

\section{Explaining Predictions of Classifiers}

We consider the problem of explaining a prediction $f(\x)$ associated to an input $\x$ by assigning to each input variable $x_d$ a score $R_d$ determining how relevant the input variable is for explaining the prediction. The scores can be pooled into groups of input variables (e.g. all \textit{word2vec} dimensions of a word, or all components of a RGB pixel), such that they can be visualized as heatmaps of highlighted texts, or as images.

\subsection{Layer-Wise Relevance Propagation}\label{LRP}

Layer-wise relevance propagation \cite{Bach} is a newly introduced technique for obtaining these explanations. It can be applied to various machine learning classifiers such as deep convolutional neural networks. The LRP technique produces a \emph{decomposition} of the function value $f(\x)$ on its input variables, that satisfies the conservation property:
\begin{equation}
f(\x)= {\textstyle \sum_d} R_d.
\label{eq:conservation}
\end{equation}
The decomposition is obtained by performing a backward pass on the network, where for each neuron, the relevance associated with it is redistributed to its predecessors. Considering neurons mapping a set of $n$ inputs $(x_i)_{i \in [1,n]}$ to the neuron activation $x_j$ through the sequence of functions:
\begin{align*}
z_{ij} &= x_i w_{ij} + {\textstyle \frac{b_j}{n}}\\
z_j &= {\textstyle \sum_i} z_{ij}\\
x_j &= g(z_j)
\end{align*}
where for convenience, the neuron bias $b_j$ has been distributed equally to each input neuron, and where $g(\cdot)$ is a monotonously increasing activation function. Denoting by $R_i$ and $R_j$ the relevance associated with $x_i$ and $x_j$, the relevance is redistributed from one layer to the other by defining messages $R_{i \leftarrow j}$ indicating how much relevance must be propagated from neuron $x_j$ to its input neuron $x_i$ in the lower layer. These messages are defined as:
$$
R_{i \leftarrow j} = \frac{z_{ij} + \frac{s(z_j)}{n}}{\sum_{i} z_{ij} + s(z_j)} R_j
$$
where $s(z_j) = \epsilon \cdot (1_{z_j \geq 0} - 1_{z_j < 0})$ is a stabilizing term that handles near-zero denominators, with $\epsilon$ set to $0.01$. The intuition behind this local relevance redistribution formula is that each input $x_i$ should be assigned relevance proportionally to its contribution in the forward pass, in a way that the relevance is preserved ($\sum_i R_{i \leftarrow j} = R_j$).

Each neuron in the lower layer receives relevance from all upper-level neurons to which it contributes
$$
R_i = {\textstyle \sum_j} R_{i \leftarrow j}.
$$
This pooling ensures layer-wise conservation: $\sum_i R_i = \sum_j R_j$. 
Finally, in a max-pooling layer, all relevance at the output of the layer is redistributed to the pooled neuron with maximum activation (i.e.\ winner-take-all). 
An implementation of LRP can be found in \cite{LapJMLR16} and downloaded from \texttt{www.heatmapping.org}\footnote{Currently the available code is targeted on image data.}.

\subsection{Sensitivity Analysis}\label{SA}

An alternative procedure called sensitivity analysis (SA) produces explanations by scoring input variables based on how they affect the decision output locally \cite{DimopoulosBL95,Gevrey2003249}. The sensitivity of an input variable is given by its squared partial derivative:
$$
R_d = \Big(\frac{\partial f}{\partial x_d} \Big)^2.
$$
Here, we note that unlike LRP, sensitivity analysis does not preserve the function value $f(\x)$, but the squared $l_2$-norm of the function gradient:
\begin{equation}\label{eq:conservation_SA}
\| \nabla_{\x} f(\x) \|_2^2 = {\textstyle \sum_d} R_d.
\end{equation}
This quantity is however not directly related to the amount of evidence for the category to detect. Similar gradient-based analyses \cite{Denil2,Li} have been recently applied in the NLP domain, and were also used by \newcite{Simonyan} in the context of image classification. While recent work uses different relevance definitions for a group of input variables (e.g. gradient magnitude in \newcite{Denil2} or \textit{max}-norm of absolute value of simple derivatives in \newcite{Simonyan}), in
the present work (unless otherwise stated) we employ the squared  $l_2$-norm of gradients allowing for decomposition of Eq.~\ref{eq:conservation_SA} as a {\it sum} over relevances of input variables.

\section{Experiments}
For the following experiments we use the \textit{20news-bydate} version of the \textit{20Newsgroups}\footnote{\texttt{http://qwone.com/\%7Ejason/20Newsgroups/}} dataset consisting of 11314/7532 train/test documents evenly distributed among twenty fine-grained categories.

\subsection{CNN Model}

As a document classifier we employ a word-based CNN similar to \newcite{Kim} consisting of the following sequence of layers:
\begin{equation*} 
\texttt{Conv} \xrightarrow{} \texttt{ReLU} \xrightarrow{} \texttt{1-Max-Pool}  \xrightarrow{}  \texttt{FC} \\   
\end{equation*}
By \texttt{1-Max-Pool} we denote a max-pooling layer where the pooling regions span the whole text length, as introduced in \cite{Collobert}.
\texttt{Conv},  \texttt{ReLU} and \texttt{FC} denote the convolutional layer, rectified linear units activation and
fully-connected linear layer.
For building the CNN numerical input we concatenate horizontally 300-dimensional pre-trained {\it word2vec}\footnote{\texttt{GoogleNews-vectors-negative300, https://code.google.com/p/word2vec/}} vectors \cite{Miko1ov1}, in the same order the corresponding words appear in the pre-processed document, and further keep this input representation fixed during training.
The convolutional operation we apply in the first neural network layer is one-dimensional and along the text sequence direction (i.e. along the horizontal direction). The receptive field of the convolutional layer neurons spans the entire word embedding space in vertical direction, and covers two consecutive words in horizontal direction. The convolutional layer filter bank contains 800 filters.

\subsection{Experimental Setup}

As pre-processing we remove the document headers, tokenize the  text with NLTK\footnote{We employ NLTK's version 3.1 recommended tokenizers \texttt{sent\_tokenize} and \texttt{word\_tokenize}, module \texttt{nltk.tokenize}.}, filter out punctuation and numbers\footnote{We retain only tokens composed of the following characters: alphabetic-character, apostrophe, hyphen and dot, and containing at least one alphabetic-character.}, and finally truncate each document to the first 400 tokens.
We train the CNN by stochastic mini-batch gradient descent with momentum (with $l_2$-norm penalty and dropout).
Our trained classifier achieves a classification accuracy of 80.19\%\footnote{To the best of our knowledge, the best published \textit{20Newsgroups} accuracy is 83.0\% \cite{Paskov}. However we notice that for simplification we use a fixed-length document representation, and our main focus is on explaining classifier decisions, not on improving the classification state-of-the-art.}. 

Due to our input representation, applying LRP or SA to our neural classifier yields one relevance value per word-embedding dimension.
From these single input variable relevances to obtain word-level relevances, we sum up the relevances over the word embedding space in case of LRP, and (unless otherwise stated) take the squared $l_2$-norm of the corresponding word gradient in case of SA. 
More precisely, given an input document $d$ consisting of a sequence $(w_1, w_2,..., w_N)$ of $N$ words, each word being represented by a $D$-dimensional word embedding, we compute the relevance $R(w_t)$ of the $t^\mathrm{th}$ word in the input document, through the summation:
\begin{equation}\label{eq:word_level}
R(w_t) = \sum_{i=1}^{D} R_{i, t}
\end{equation}
where $R_{i, t}$ denotes the relevance of the input variable corresponding to the $i^\mathrm{th}$ dimension of the $t^\mathrm{th}$ word embedding, obtained by LRP or SA as specified in Sections~\ref{LRP}~\&~\ref{SA}.

In particular, in case of SA, the above word relevance can equivalently be expressed as:
\begin{equation}\label{eq:word_level_SA}
R_{\mathrm{SA}}(w_t) = \| \nabla_{{w_t}} f({d}) \|_2^2 
\end{equation}
where $f({d})$ represents the classifier's prediction for document $d$.

Note that the resulting LRP word relevance is signed, while the SA word relevance is positive.

In all experiments, we use the term {\it target} class to identify the function $f(x)$  to analyze in the relevance decomposition. This function maps the neural network input to the neural network output variable corresponding to the target class.

\subsection{Evaluating Word-Level Relevances}

In order to evaluate different relevance models, we perform a sequence of ``word deletions'' (hereby for deleting a word we simply set the word-vector to zero in the input document representation), and track the impact of these deletions on the classification performance.
We carry out two deletion experiments, starting either with the set of test documents that are initially classified correctly, or with those that are initially classified wrongly\footnote{For the deletion experiments we consider only the test documents whose pre-processed length is greater or equal to 100 tokens, this amounts to a total of 4963 documents.}.
We estimate the LRP/SA word relevances using as target class the true document class. Subsequently we delete words in decreasing resp.\ increasing order of the obtained word relevances.

\begin{figure}
\centering
\includegraphics[width=0.95\columnwidth]{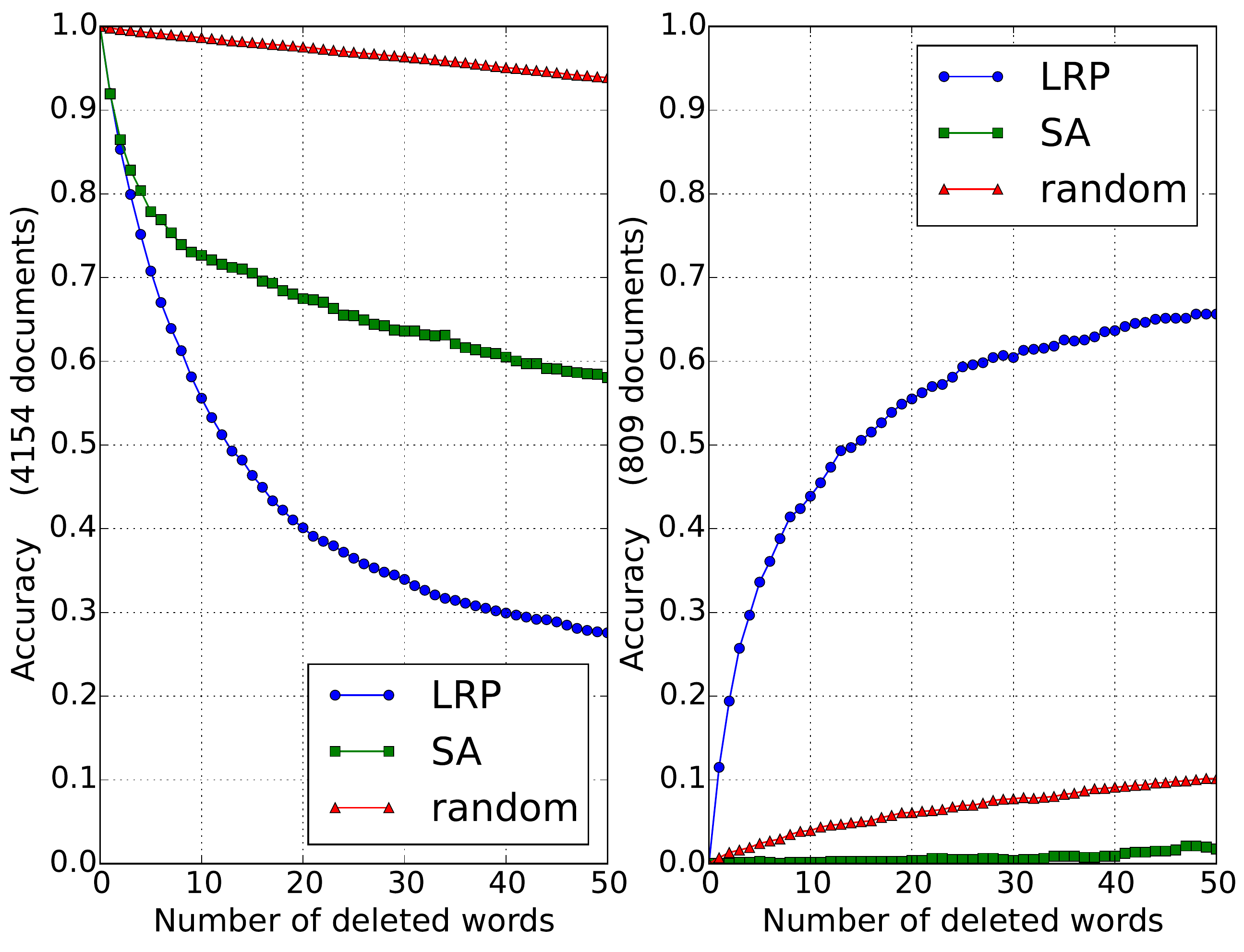}
\caption{Word deletion on initially correct (left) and false (right) classified test documents, using either LRP or SA. The target class is the true document class, words are deleted in decreasing (left) and increasing (right) order of their relevance. Random deletion is averaged over 10 runs (std $<$ 0.0141). A steep decline (left) and incline (right) indicate informative word relevances.}\label{fig:deletion}
\end{figure}

Fig.~\ref{fig:deletion} summarizes our results.
We find that LRP yields the best results in both deletion experiments. Thereby we provide evidence that LRP positive relevance is targeted to words that {\it support} a classification decision, while LRP negative relevance is tuned upon words that {\it inhibit} this decision.
In the first experiment the SA classification accuracy curve decreases significantly faster than the random curve representing the performance change when randomly deleting words, indicating that SA is able to identify relevant words. However, the SA curve is clearly above the LRP curve indicating that LRP provides better explanations for the CNN predictions. Similar results have been reported for image classification tasks \cite{SamArXiv15}.
The second experiment indicates that the classification performance increases when deleting words with the lowest LRP relevance, while small SA values points to words that have less influence on the classification performance than random word selection. This result can partly be explained by the fact that in contrast to SA, LRP provides {\it signed} explanations.
More generally the different quality of the explanations provided by SA and LRP can be attributed to their different objectives: while LRP aims at decomposing the {\it global} amount of evidence for a class $f(x)$, SA is build solely upon derivatives and as such describes the effect of local {\it variations} of the input variables on the classifier decision. 
For a more detailed view of SA,  
as well as an interpretation of the LRP propagation rules as a {\it deep Taylor decomposition} see \newcite{MontavonArXiv15}.

\subsection{Document Highlighting}

\begin{figure*}
\centering
\includegraphics[clip=true, trim=23mm 9.1cm 25mm 1.6cm, width=\textwidth, resolution=300]{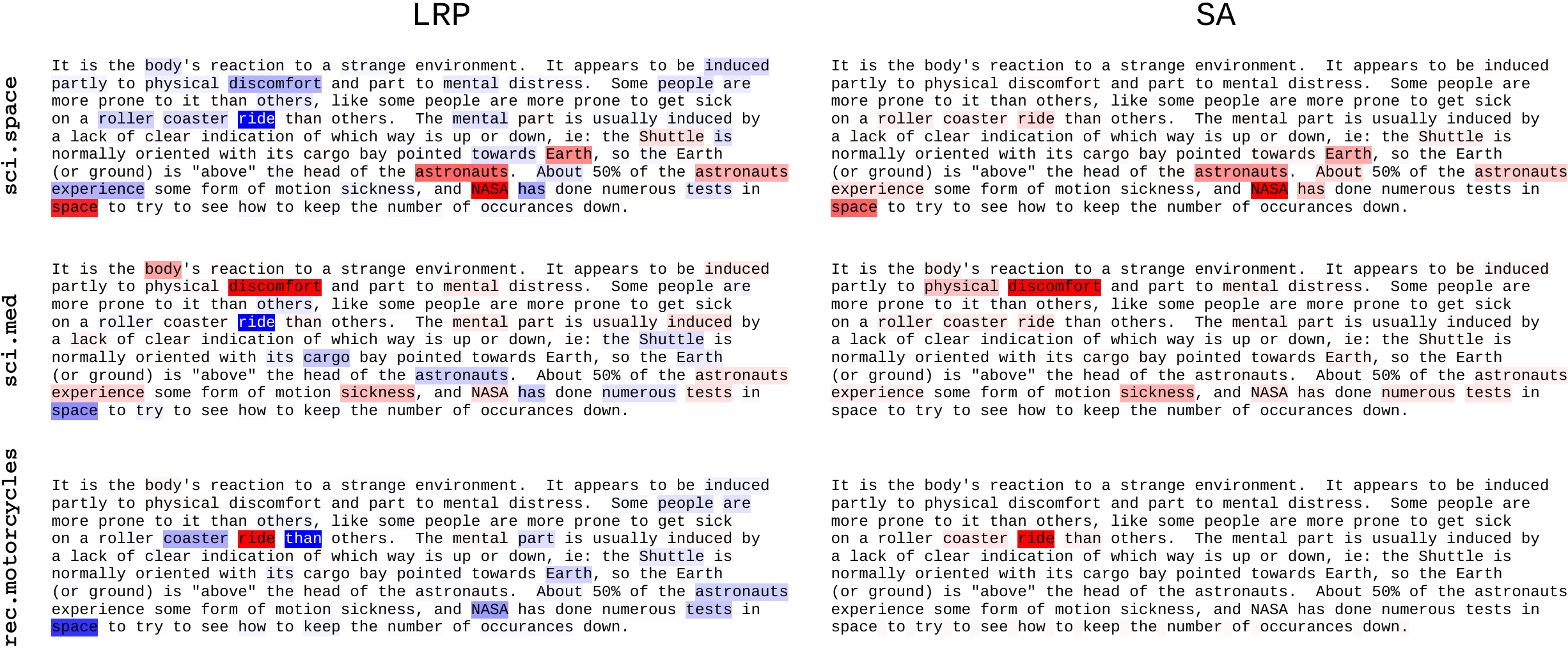}
\caption{Heatmaps for the test document {\tt sci.space 61393} (correctly classified), using either layer-wise relevance propagation (LRP) or sensitivity analysis (SA) for highlighting words. Positive relevance is mapped to red, negative to blue. The target class for the LRP/SA explanation is indicated on the left.}\label{fig:heatmap}
\end{figure*}

Word-level relevances can be used for highlighting purposes. In Fig.~\ref{fig:heatmap} we provide such visualizations on one test document for different relevance target classes, using either LRP or SA relevance models.
We can observe that while the word {\tt ride} is highly negative-relevant for LRP when the target class is not {\tt rec.motorcycles}, it is positively highlighted (even though not heavily) by SA. This suggests that SA does not clearly discriminate between words speaking {\it for} or {\it against} a specific classifier decision, while LRP is more discerning in this respect.

\subsection{Document Visualization}

\textit{Word2vec} embeddings are known to exhibit linear regularities representing semantic relationships between words \cite{Miko1ov1}.
We explore if these regularities can be transferred to a document representation, when using as a document vector a linear combination of {\it word2vec} embeddings.
As a weighting scheme we employ LRP or SA scores, with the classifier's predicted class as the target class for the relevance estimation. For comparison we perform uniform weighting, where we simply sum up the word embeddings of the document words (SUM). 

For SA we use either the $l_2$-norm or squared $l_2$-norm for pooling word gradient values along the \textit{word2vec} dimensions, i.e. in addition to the standard SA word relevance defined in Eq.~\ref{eq:word_level_SA}, we use as an alternative $R_{\mathrm{SA}(l_2)}(w_t) = \| \nabla_{{w_t}} f({d}) \|_2$ and denote this relevance model by SA$(l_2)$.

\begin{figure*}
\centering
\includegraphics[width=\textwidth]{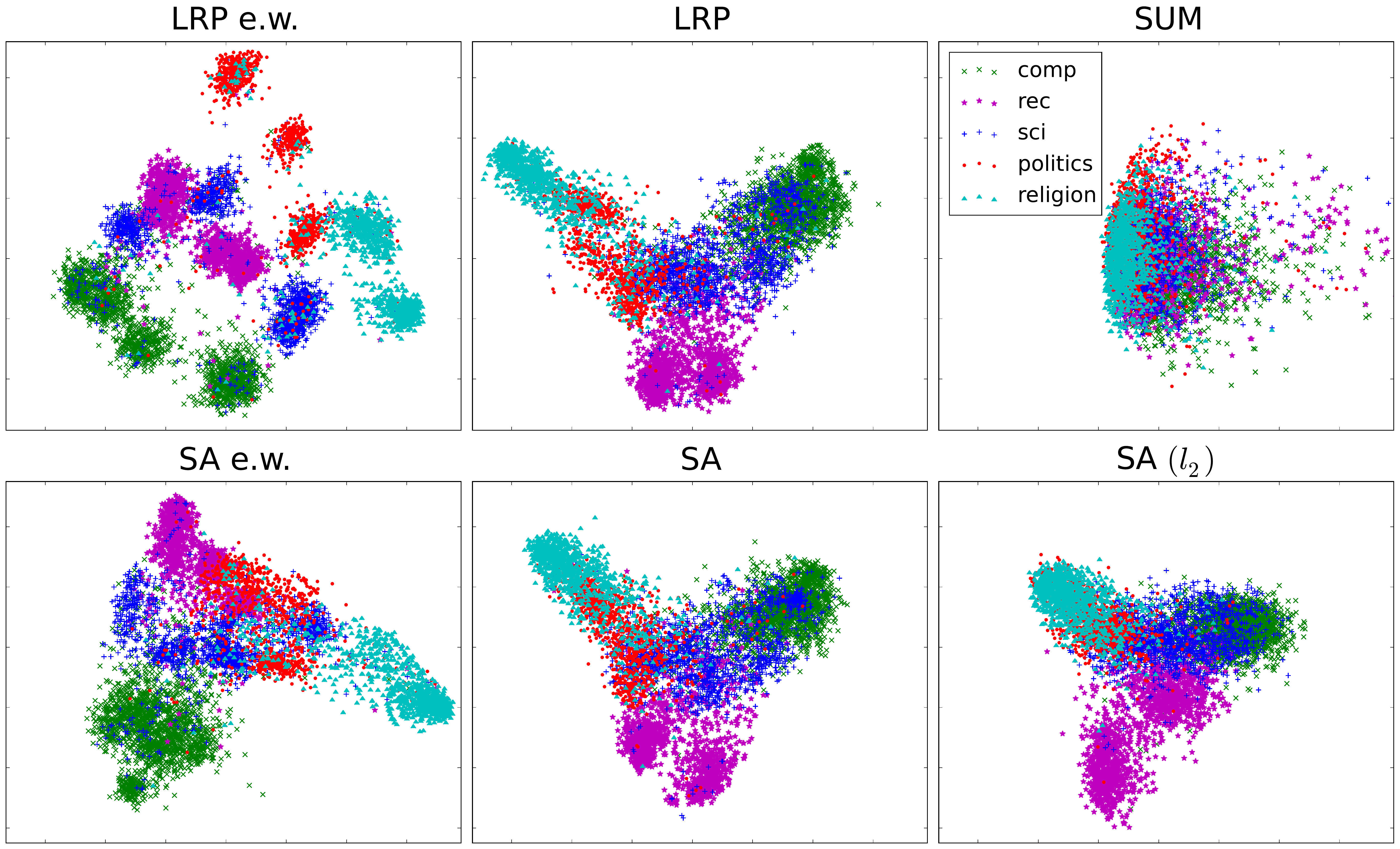}
\caption{PCA projection of the {\it 20Newsgroups} test documents formed by linearly combining {\it word2vec} embeddings. The weighting scheme is based on word-level relevances, or on single input variable relevances (e.w.), or uniform (SUM). The target class for relevance estimation is the predicted document class. SA$(l_2)$ corresponds to a variant of SA with simple $l_2$-norm pooling of word gradient values. All visualizations are provided on the same equal axis scale.}\label{fig:PCA}
\end{figure*}

For both LRP and SA, we employ different variations of the weighting scheme. More precisely, given an input document $d$ composed of the sequence $(w_1, w_2,..., w_N)$ of $D$-dimensional \textit{word2vec} embeddings, we build new document representations $d'$ and $d'_{\mathrm{e.w.}}$\footnote{The subscript {$\mathrm{e.w.}$} stands for {\it element-wise}.} by either using word-level relevances $R(w_t)$ (as in Eq.~\ref{eq:word_level}), or through element-wise multiplication of word embeddings with single input variable relevances $(R_{i, t})_{i \in [1,D]}$ (we recall that $R_{i, t}$ is the relevance of the input variable corresponding to the $i^{\mathrm{th}}$ dimension of the $t^{\mathrm{th}}$ word in the input document $d$).
More formally we use:
\begin{equation*}\label{eq:document_model}
d'          \; = \;   \sum_{t=1}^{N} \; {{R(w_t)} \cdot {w_t}}
\end{equation*}
or
\begin{equation*}\label{eq:document_model_ew}
d'_{\mathrm{e.w.}}   \; = \;   \sum_{t=1}^{N} \; { \begin{bmatrix} R_{1, t} \\ R_{2, t} \\ \vdots \\ R_{D, t} \end{bmatrix}  \odot {w_t}}
\end{equation*}
where $\odot$ is an element-wise multiplication.
Finally we normalize the document vectors $d'$ resp. $d'_{\mathrm{e.w.}}$ to unit $l_2$-norm  and perform a PCA projection. In Fig.~\ref{fig:PCA} we label the resulting 2D-projected test documents using five top-level document categories. 

For word-based models $d'$, we observe that while standard SA and LRP both provide similar visualization quality, the SA variant with simple $l_2$-norm yields partly overlapping and dense clusters, still all schemes are better than uniform\footnote{We also performed a TFIDF weighting of word embeddings, the resulting 2D-visualization was very similar to uniform weighting (SUM).} weighting.
In case of SA note that, even though the power to which word gradient norms are raised ($l_2$ or $l_2^2$) affects the present visualization experiment, it has no influence on the earlier described ``word deletion'' analysis.

For element-wise models $d'_{\mathrm{e.w.}}$, we observe slightly better separated clusters for SA, and a clear-cut cluster structure for LRP.

\section{Conclusion}
Through word deleting we quantitatively evaluated and compared two classifier explanation models, and pinpointed LRP to be more effective than SA.
We investigated the application of word-level relevance information for document highlighting and visualization.
We derive from our empirical analysis that the superiority of LRP stems from the fact that it reliably not only links to determinant words that {\it support} a specific classification decision, but further distinguishes, within the preeminent words, those that are {\it opposed} to that decision. 

Future work would include applying LRP to other neural network architectures (e.g. character-based or recurrent models) on further NLP tasks, as well as exploring how relevance information could be taken into account to improve the classifier's training procedure or prediction performance.

\section*{Acknowledgments}
This work was supported by the German Ministry for Education and Research as Berlin Big Data Center BBDC (01IS14013A) and the Brain Korea 21 Plus Program through the National Research Foundation of Korea funded by the Ministry of Education.

\bibliography{acl2016}
\bibliographystyle{acl2016}

\appendix
\end{document}